\documentclass{article}
\PassOptionsToPackage{}{natbib}  

\usepackage{conf/colm2024_conference}
\colmfinalcopy  

\usepackage{color}
\usepackage{xcolor}

\definecolor{darkblue}{rgb}{0, 0, 0.5}
\definecolor{darkgreen}{rgb}{0, 0.5, 0}

\usepackage{microtype}
\usepackage{graphicx}
\usepackage{subcaption}
\usepackage{booktabs}
\usepackage[
    colorlinks,
    linkcolor=darkblue,
    citecolor=darkblue,
    linkcolor=darkblue,
    urlcolor=darkblue
]{hyperref}
\usepackage{amsmath}
\usepackage{amssymb}
\usepackage{mathtools}
\usepackage{amsthm}
\usepackage[capitalize,noabbrev]{cleveref}

\usepackage[utf8]{inputenc}
\usepackage[T1]{fontenc}
\usepackage{url}
\usepackage{amsfonts}
\usepackage{nicefrac}
\usepackage{latexsym}
\usepackage{array}
\usepackage{multirow}
\usepackage{alltt}
\usepackage{enumerate}
\usepackage{bbm}
\usepackage{eucal}
\usepackage{xspace}
\usepackage{stmaryrd}
\usepackage{bussproofs}
\usepackage{mathpartir}
\usepackage{tikz}
\usepackage{pgfmath}
\usepackage{xurl}
\usepackage{mdframed}
\usepackage{enumitem}
\usepackage{tabularx}
\usepackage{listings}
\usepackage{xparse}
\usepackage{xifthen}
\usepackage{multicol}
\usepackage{xfrac}
\usepackage{appendix}
\usepackage{etoolbox}
\usepackage{floatflt}
\usepackage{wrapfig}
\usepackage{tablefootnote}
\usepackage{makecell}
\usepackage{verbatim}
\usepackage{setspace}
\usepackage{floatrow}

\usepackage{algorithm}
\usepackage[%
    noEnd=true,%
    indLines=true,%
    italicComments=false,%
    commentColor=darkgreen,%
]{algpseudocodex}

\theoremstyle{definition}

\theoremstyle{remark}

\makeatletter
\newcommand{\verbatimfont}[1]{\def\verbatim@font{#1}}%
\makeatother

\newcommand{\sample}[1]{
    \verbatimfont{\rmfamily}
    \begin{quote}
        \verbatiminput{#1}
    \end{quote}
}

\makeatletter
\patchcmd{\l@section}
  {\addvspace{1.0em \@plus\p@}}
  {}
  {}{}
\makeatother

\newcommand{\cuemail}[1]{\href{#1@cornell.edu}{#1}}

\renewcommand{\toprule}{\noalign{\hrule height 0.9pt\vspace{0.15cm}}}
\renewcommand{\midrule}{\noalign{\vspace{0.1cm}\hrule height 0.7pt\vspace{0.1cm}}}
\renewcommand{\bottomrule}{\noalign{\vspace{0.1cm}\hrule height 0.9pt\vspace{0.1cm}}}

\newcommand{\tablerule}{\noalign{\vspace{0.1cm}\hrule height 0.35pt\vspace{0.1cm}}}

\newcommand{\multirowrotate}[3]{\parbox[t]{2mm}{\multirow{#1}{*}{\rotatebox[origin=c]{#2}{#3}}}}

\newcommand{\bigo}{\mathcal{O}}

\newcommand{\I}{\mathrm{I}}

\makeatletter
\DeclareDocumentCommand{\R}{O{} O{}}{
    \ifx&#1&\@empty\relax {
        \ifx&#2&\@empty\relax {
            \mathbb{R}
        } \else {
            \mathbb{R}^{#2}
        }
        \fi
    } \else {
        \ifx&#2&\@empty\relax {
            \mathbb{R}^{#1}
        } \else {
            \mathbb{R}^{#1 \times #2}
        }
        \fi
    } 
    \fi
}
\makeatother

\DeclareMathOperator{\dd}{d}
\newcommand{\ddt}[1]{\frac{{\dd}#1}{{\dd}t}}

\newcommand{\discrete}[1]{
    \vbox{%
        \hrule height 0.55pt  
        \kern0.3ex  
        \hbox{%
            \kern-0.0em  
            \ifmmode#1\else\ensuremath{#1}\fi  
            \kern-0.0em  
        }
    }
}

\DeclareMathOperator{\A}{\mathrm{A}}
\DeclareMathOperator{\B}{\mathrm{B}}
\DeclareMathOperator{\C}{\mathrm{C}}
\DeclareMathOperator{\D}{\mathrm{D}}

\newcommand{\linear}[2]{W_{#1}#2}
\DeclareMathOperator{\softplus}{softplus}

\DeclareMathOperator{\bpb}{BPB}
\DeclareMathOperator{\ppl}{PPL}

\DeclareSymbolFont{matha}{OML}{txmi}{m}{it}
\DeclareMathSymbol{\varv}{\mathord}{matha}{118}

\newcommand{\ssmname}{MambaByte\xspace}

\newcommand{\smallssmlayers}{$53$}
\newcommand{\smallssmsize}{$353$M}
\newcommand{\smallssmdim}{$1,024$}
\newcommand{\smallssmexp}{$2$}
\newcommand{\smallssm}{\ssmname-\smallssmsize\xspace}
\newcommand{\smallssmconfig}{%
    $n = \text{\smallssmlayers}$,
    $d = \text{\smallssmdim}$,
    $e = \text{\smallssmexp}$%
}

\newcommand{\mediumssmsize}{$972$M}
\newcommand{\mediumssmdim}{$1,792$}
\newcommand{\mediumssmexp}{$2$}
\newcommand{\mediumssmlayers}{$48$}
\newcommand{\mediumssm}{\ssmname-\mediumssmsize\xspace}
\newcommand{\mediumssmconfig}{%
    $n = \text{\mediumssmlayers}$,
    $d = \text{\mediumssmdim}$,
    $e = \text{\mediumssmexp}$%
}

\newcommand{\bigssmsize}{$1.6$B}
\newcommand{\bigssmdim}{$2,304$}
\newcommand{\bigssmexp}{$2$}
\newcommand{\bigssmlayers}{$48$}
\newcommand{\bigssm}{\ssmname-\bigssmsize\xspace}

\newcommand{\megabytesmall}{MegaByte-$758$M+$262$M\xspace}
\newcommand{\megabytemedium}{MegaByte-$1.3$B+$218$M\xspace}
\newcommand{\megabytelarge}{MegaByte-$1.3$B+$350$M\xspace}

\let\svthefootnote\thefootnote
\newcommand\freefootnote[1]{%
  \let\thefootnote\relax%
  \footnotetext{#1}%
  \let\thefootnote\svthefootnote%
}

\algnewcommand{\LeftComment}[1]{~~~~\textcolor{darkgreen}{\(\triangleright\) #1}}

\newfloatcommand{capbtabbox}{table}[][\FBwidth]

\title{\ssmname: Token-free Selective State Space Model}

\author{%
    Junxiong~Wang, 
    Tushaar~Gangavarapu,
    Jing~Nathan~Yan,
    Alexander~M.~Rush \\[3pt]
    Cornell University \\
    \texttt{\{%
        \cuemail{jw2544},%
        \cuemail{tg352},%
        \cuemail{jy858},%
        \cuemail{arush}%
    \}@cornell.edu}
    \ifcolmpreprint%
        \\[3pt]
        (Version~$2$)
    \fi
}

\begin{document}
\maketitle

\ifcolmpreprint%
    \freefootnote{$^0$Compared to V$1$, this draft includes the (subword) Mamba baseline, experimental results on length extrapolation and synthetic noise, and an adaptation of speculative decoding for byte-level models.}
\fi

\addtocontents{toc}{\protect\setcounter{tocdepth}{-10}}
\begin{abstract}
    Token-free language models learn directly from raw bytes and remove the inductive bias of subword tokenization. Operating on bytes, however, results in significantly longer sequences. In this setting, standard autoregressive Transformers scale poorly as the effective memory required grows with sequence length. The recent Mamba state space model (SSM) development offers an appealing alternative approach with a fixed-sized memory state and efficient decoding. We propose \ssmname, a token-free adaptation of the Mamba SSM trained autoregressively on byte sequences. In terms of modeling, we show \ssmname to be competitive with, and even to outperform, state-of-the-art subword Transformers on language modeling tasks while maintaining the benefits of token-free language models, such as robustness to noise. In terms of efficiency, we develop an adaptation of speculative decoding with tokenized drafting and byte-level verification. This results in a $2.6\times$ inference speedup to the standard MambaByte implementation, showing similar decoding efficiency as the subword Mamba. These findings establish the viability of SSMs in enabling token-free language modeling.

\end{abstract}
\section{Introduction}
\label{sec:intro}

When defining a language model, a base tokenization is typically used---either words \citep{bengio2000neural}, subwords \citep{schuster2012japanese,sennrich2015neural,wu2016google,wang2020neural}, or characters \citep{gao2020character}. Of these, subword tokenization has been the most popular choice, as it achieves a natural compromise between training efficiency and the ability to handle out-of-vocabulary words. However, several works, e.g., \citet{xue2022byt5}, have noted issues with subword tokenizers, such as a lack of robustness to typos, spelling and capitalization variations, and morphological changes.


Modeling byte sequences, i.e., mapping from raw data to predictions without any intermediate tokenization, offers an alternative approach with less inductive bias \citep{choe2019bridging,alrfou2019character,clark2022canine,tay2022charformer,xue2022byt5, yu2023megabyte}. Compared to subword models, byte-level language models can generalize more easily across orthographic and morphological variants. Of course, modeling text as bytes means that the resultant sequences are significantly longer than their subword counterparts. This change pushes the modeling and efficiency issues upstream into the architecture itself.

\begin{figure*}[t]
    \centering
    \includegraphics[width=1\textwidth]{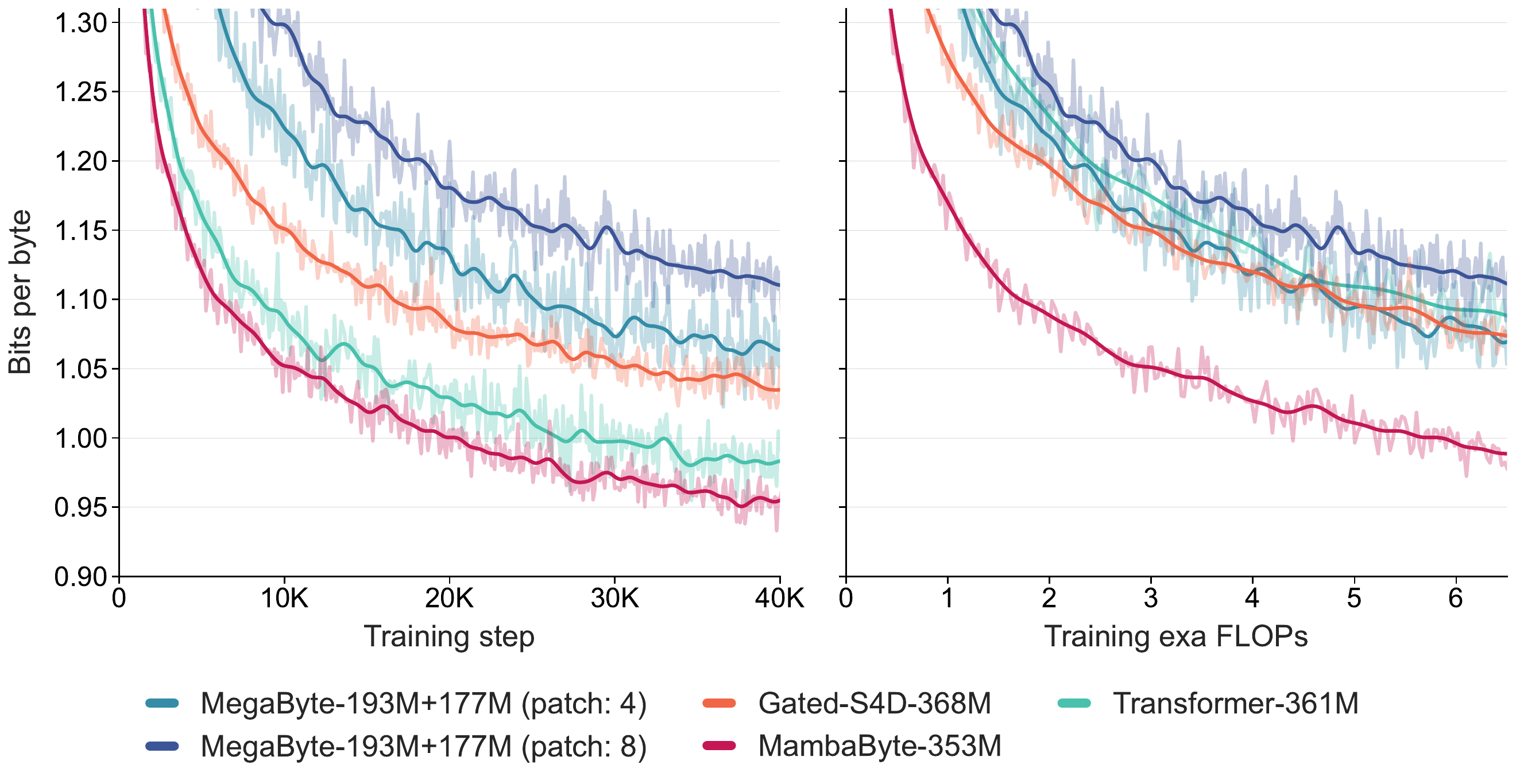}
    \caption{\textbf{Benchmarking byte-level models with a fixed parameter budget.} 
    Language modeling results on PG19 ($8,192$ consecutive bytes), comparing the standard Transformer \citep{vaswani2017attention,su2021roformer}, MegaByte Transformer \citep{yu2023megabyte}, gated diagonalized S4 \citep{mehta2023long}, and MambaByte.
    (Left) Model loss over training step. 
    (Right) FLOP-normalized training cost.
    \ssmname reaches Transformer loss in less than one-third of the compute budget.}
    \label{fig:pg19-test-performance}
\end{figure*}

These issues are particularly pronounced for autoregressive Transformers \citep{vaswani2017attention}, which dominate language modeling \citep{brown2020language,touvron2023llama2}. Due to the quadratic nature of attention, Transformer efficiency scales poorly for long (byte) sequences \citep{zhang2022opt}. Researchers have \textit{compressed} the internal Transformer representation to work with long sequences, for instance, developing length-aware modeling approaches \citep{dai2020funnel,nawrot2022hierarchical}, where groups of tokens are merged within the intermediate layers. The MegaByte Transformer \citep{yu2023megabyte} is of particular relevance, which uses compression in the form of fixed-size patches of bytes as a subword analog in combination with a byte-level decoder. These methods lower computational costs\footnote{However, our experiments (see Figure~\ref{fig:pg19-test-performance}) indicate that patching can also lower the model performance compared to the standard Transformer.} but change the modeling behavior to match the data.

In this work, we propose \ssmname, a byte-level language model without representational compression. The model is an application of the recently introduced Mamba architecture \citep{gu2023mamba}. Mamba builds off the approach pioneered by state space models (SSMs) \citep{gu2021efficiently,gupta2022diagonal,gu2022parameterization,smith2023simplified,fu2022hungry} by introducing a selection mechanism that has been shown to be nearly as effective as Transformers for discrete data. Our key observation is that, unlike Transformers, Mamba has a (large) fixed-sized memory state that is independent of context length, roughly analogous to a large recurrent neural network hidden state. This naturally removes a major modeling and efficiency issue for byte-level language modeling without requiring specialized architectures such as global patching.

Even with effective training, byte-level models still suffer from the challenge of efficient decoding, as generating one character at a time requires running the language model in serial one byte at a time. To improve the inference efficiency, we propose an adaptation of speculative decoding \citep{leviathan2023fast,chen2023accelerating,xia2023speculative} to byte-level models. The approach uses a fast subword model for autoregressive drafting, followed by byte-level verification. While this approach could be applied to any byte-level model, it is particularly efficient for SSM-style models since the byte-level verification step can use the same parallel scan code path that makes these models efficient to train.

Experiments compare \ssmname to Transformers, SSMs, and MegaByte (patching) architectures in a fixed parameter and fixed compute setting on several long-form language modeling datasets. Figure~\ref{fig:pg19-test-performance} summarizes our main findings. Compared to byte-level Transformers, \ssmname achieves better performance faster and is significantly more compute-efficient. We also compare \ssmname with tokenized subword baselines using Transformers and SSMs, and find that \ssmname is competitive in loss while also demonstrating improved robustness in handling subword noise, such as input text corruptions. Through our speculative subword drafting and byte-level verification approach, we show that \ssmname can be run as fast as the subword Mamba for text generation. We believe these results validate the potential for tokenizer-free models as a practical alternative to subword Transformers for language modeling.



\section{State space models and the Mamba architecture}
\label{sec:model}





\paragraph{Method: Selective SSMs.} SSMs model the evolution of a hidden state across time through a first-order differential equation. Linear time-invariant SSMs \citep{gu2021efficiently,gupta2022diagonal,gu2022parameterization,smith2023simplified} have shown promising results in deep learning across several modalities. However, \citet{gu2023mamba} have recently argued that the constant dynamics of these approaches lack input-dependent context \textit{selection} in the hidden state, which may be necessary for tasks such as language modeling. To this end, they define the time-varying continuous state dynamics for a given input $x(t) \in \R$, hidden state $h(t) \in \R[n]$, and output $y(t) \in \R$ at time $t$ as:
\begin{align}
\label{eq:mamba}
    \ddt{h(t)} = {\A}h(t) + \B(t) x(t); \quad y(t) = \C(t) h(t),
\end{align}
which is parameterized by a diagonal time-invariant system matrix $\A \in \R[n][n]$ and time-dependent input and output matrices $\B(t) \in \R[n][1]$ and $\C(t) \in \R[1][n]$.



To model discrete-time sequences, the continuous-time dynamics in (\ref{eq:mamba}) must be approximated through discretization. This results in a discrete-time hidden state recurrence with new matrices at each timestep, $\discrete{\A}$, $\discrete{\B}$, and $\discrete{\C}$, such that
\begin{align}
\label{eq:discrete-ssm}
    h[k] = \discrete{\A}[k] h[k-1] + \discrete{\B}[k] x[k]; \quad y[k] = \discrete{\C}[k] h[k].
\end{align}

Observe that (\ref{eq:discrete-ssm}) resembles a linear version of a recurrent neural network and can be applied in this recurrent form during language model generation. The discretization requires a timestep, $\Delta[k]$, for each input position, corresponding to treating $x[k] = x\left(t_k\right)$ for $t_k =\sum_{j=1}^{k} \Delta[j]$. The discrete-time matrices $\discrete{\A}$, $\discrete{\B}$, and $\discrete{\C}$ can then be computed from $\Delta[k]$.

\paragraph{Architecture: Mamba.} In Mamba, the SSM terms are input-selective, i.e., $\B$, $\C$, and $\Delta$ are defined as functions of the input $x[k] \in \R[d]$:
\begin{align}
\label{eq:selective-ssm}
    \Delta[k] = \softplus(\linear{\Delta}{(\linear{R}{x}[k]})); \quad \B(t_k) = \linear{\B}{x}[k], 
\end{align}
where $W_{\B} \in \R[n][d]$ ($\C$ is similarly defined), $W_{\Delta} \in \R[d][r]$ and $W_{R} \in \R[r][d]$ (for some $r \ll d$) are learnable weights, and softplus ensures positivity. Note that the SSM parameters $\A$, $\B$, and $\C$ are identical for each input dimension $d$, but the timesteps $\Delta$ are distinct; this results in a hidden state size of $n \times d$ per timestep $k$. (See Appendix~\ref{app:selection} for an illustration of how Mamba models discrete sequences and other specifics on discretization and selectivity.)

Mamba embeds this SSM layer into a full neural network language model. Specifically, the model utilizes a stack of gated layers inspired by the previous gated SSM \citep{mehta2023long}. Figure~\ref{fig:model-arch} (right) in Appendix~\ref{app:compute} shows the Mamba architecture combining the SSM layer with a gated neural network.

\paragraph{Implementation: Parallel scans for linear recurrences.} At training time, we have access to the entire sequence $x$, allowing us to compute the linear recurrence more efficiently. \citet{smith2023simplified} demonstrated the use of work-efficient parallel scans \citep{blelloch1990prefix} for efficiently computing the sequential recurrence in linear SSMs. For Mamba, we first map the recurrence to a sequence of $L$ tuples, with $e_k = (A_k, b_k) \coloneqq (\discrete{\A}[k], \discrete{\B}[k]x[k])$, then define an associative operator $\bullet$ such that $e_j \bullet e_k = (A_kA_j, A_kb_j + b_k)$. Finally, we apply a parallel scan to compute the sequence $[(\discrete{\A}[1], h[1]), (\discrete{\A}[2]\discrete{\A}[1], h[2]), \ldots]$. In general, this requires $\bigo(T_\bullet \log_2(L))$ time, using $L/2$ processors, where $T_\bullet$ is the cost of a matrix-matrix multiplication. Noting $\discrete{\A}$ to be a diagonal matrix, the linear recurrence can be computed parallelly in $\bigo(n \log_2(L))$ time and $\bigo(nL)$ space. A parallel scan with a diagonal matrix is also  efficient in operation, requiring $\bigo(nL)$ FLOPs.

\section{Method}
\label{sec:method}

\paragraph{Modeling long byte-sequences.} \ssmname is an application of the Mamba architecture to byte-level language modeling. Our main observation is that unlike Transformers, whose memory scales linearly in sequence length, Mamba maintains a large fixed-size memory state, which makes it suitable for direct byte-level modeling. Formally, an $m$-layer Mamba model, each with a hidden state $h(t) \in \R[n_\text{state}][d]$, efficiently maintains and evolves a memory of $m \times n_\text{state} \times d$ floats. Noting that the Mamba hidden state memory size is independent of input context length, $L_\text{ctx}$, processing subword sequences or byte sequences requires the underlying model to compress roughly $L_\text{ctx}$ bytes in its fixed hidden state memory, irrespective of the input representation. In all but extreme cases, $m \times n_\text{state} \times d \gg L_\text{ctx}$, leaving enough space of a hidden state $h(t)$ to encode $L_\text{ctx}$ information. Therefore, if Mamba can be used for tokenized models, \ssmname should enable modeling byte-level sequences without the need for length-compression trade-offs \citep{dai2020funnel,nawrot2022hierarchical,yu2023megabyte}.


Utilizing a fixed-sized memory representation may also help avoid quadratic dependencies and improve generalization. While Transformers are designed to capture long-range dependencies, researchers have noted that the sheer number of potential interactions in a long byte-level sequence can dilute the model's focus, making it challenging to capture crucial dependencies amid a vast number of less relevant ones \citep{tworkowski2024focused}. Bytes level information is much more granular, thus necessitating the model to learn from a much larger context to make meaningful predictions.
 
Finally, training Mamba for long byte-sequences has an inherent computation benefit at scale. The computational cost for Mamba at training is $\bigo(L_\text{ctx})$, while even compressed models such as MegaByte \citep{yu2023megabyte} have a complexity of $\bigo(L_\text{ctx}^2/p^2 + L_\text{ctx}p)$ for a patch size $p$. Even with a large patch size of $L_\text{ctx}^{1/3}$, the resulting complexity is $\bigo(L_\text{ctx}^{4/3})$.



\begin{figure}[!t]
    \centering
    \includegraphics[width=.95\textwidth]{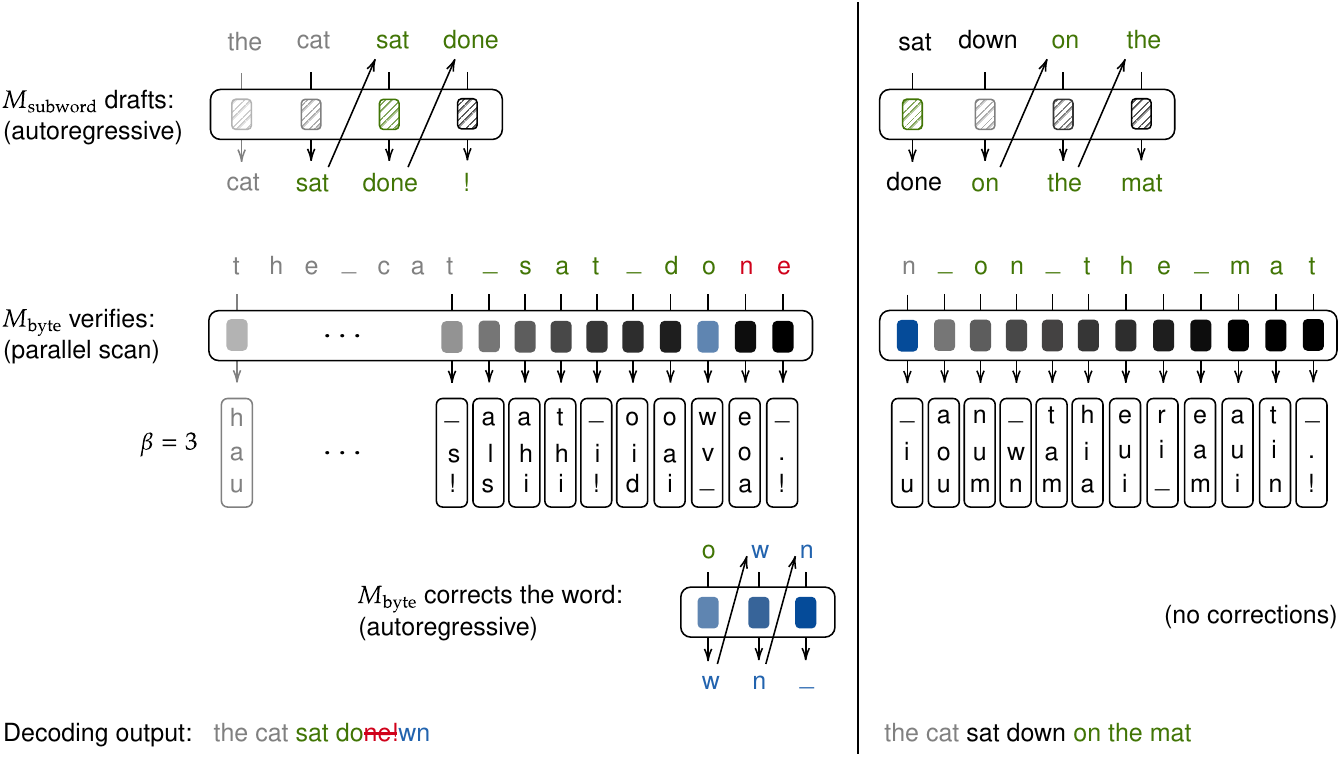}
    \caption{\textbf{Speculative decoding through subword drafting and byte-level verification.} 
    The \textcolor[HTML]{417505}{green} subwords are suggestions made by the smaller subword (Mamba) model $M_\text{subword}$, whose associated bytes fell in the top-$\beta$ autoregressive candidates of the byte-level verifier (\ssmname) model $M_\text{byte}$. The \textcolor[HTML]{D0021B}{red} and \textcolor[HTML]{2263AE}{blue} bytes are the rejections and corresponding corrections made by the verifier model. (Two steps shown using the prompt: ``the cat''.)
    }
    \label{fig:speculative-expl}
\end{figure}


\paragraph{Speculative decoding through subword drafting.} While \ssmname is computationally efficient at training, it encounters challenges in decoding, primarily because each byte is processed sequentially. To mitigate this sequential bottleneck, we propose a adaptation of speculative decoding through subword drafting and byte-level verification. Our observation is that most inference steps do not require the granularity of byte-level decoding and can benefit from faster subword drafting. Consequently, we can train token-free models, which are known to be robust to noise, and simulate subword-model-like generation, which is significantly faster. We decompose every decoding iteration into two steps: \textit{draft} using a smaller subword (Mamba) model, then \textit{verify and correct} using a larger byte-level (\ssmname) model, as illustrated in Figure~\ref{fig:speculative-expl}.

The subword Mamba model, $M_\text{subword}$, drafts $m$ subwords autoregressively while recording the associated hidden states at each timestep. The drafted subwords are converted to bytes, fed to the byte-level \ssmname model, $M_\text{byte}$, and verified using a parallel scan. We then find the bifurcation byte position $c$, the furthest position in the byte sequence verified to be in the top-$\beta$ autoregressive candidates of $M_\text{byte}$. We also find the subword bifurcation position associated with $c$, i.e., the largest position of the subword whose bytes are all verified to be correct. Drafted bytes after position $c$ are discarded. Noting that the drafter is tokenized, while the verifier is token-free, we cannot just correct for $b_{c+1}$, i.e., one byte after the bifurcation position, and continue drafting---this causes issues with drafting, especially if the tokenizer cannot find the newly updated partial subword in its pre-trained vocabulary. To avoid the possibility of the corrected partial subword being marked as out-of-vocabulary, we use the verifier model to generate bytes autoregressively until a boundary byte (e.g., space) is generated. The final decoded tokens include the verified drafted subwords and the corrected subword generated by the byte-level model. We cache the final hidden state from the \ssmname verifier and the bifurcation hidden state from the subword Mamba model for the next iteration. For completeness, we provide the algorithm for speculative decoding through subword drafting in Appendix~\ref{app:spec}.

To enable resuming during the parallel scan, we extended the fast CUDA kernel from Mamba \citep{gu2023mamba}, allowing verification to restart from the mismatched position instead of beginning from the start.

\section{Experimental setup}
\label{sec:setup}

\begin{wraptable}{r}{0.48\textwidth}
\vspace{-1em}
    \centering
    \bgroup
    \def\arraystretch{1.07}
    \resizebox{\textwidth}{!}{
    \begin{tabular}{l@{\hspace{1.2\tabcolsep}}l@{\hspace{0.5\tabcolsep}}c}
    \toprule
        Expt & Models & \makecell{FLOPs per\\train byte} \tabularnewline
    \midrule
        \multirow{2}{*}{\shortstack[l]{Medium-\\scale}} & \megabytesmall$:$ & $1.02:1$ \tabularnewline
         & \smallssm &  \tabularnewline
    \tablerule
        \multirow{2}{*}{\shortstack[l]{Large-\\scale}} & \megabytelarge$:$ & $0.54:1$ \tabularnewline
         & \mediumssm &  \tabularnewline
    \tablerule
        \multirow{2}{*}{\shortstack[l]{}} & \megabytemedium$:$ & $0.40:1$ \tabularnewline
        & \mediumssm &  \tabularnewline
    \bottomrule
    \end{tabular}}
    \caption{\textbf{Relative training FLOPs by model size.} 
    MegaByte models use a patch size of $8$.}
    \label{tab:model-flops}
    \egroup
\end{wraptable}

Our experiments compare \ssmname to a range of other tokenizer-based and token-free Transformers and SSMs. All our models employ the same training recipes. We utilize a set of diverse long-form text datasets: PG19 \citep{rae2020compressive}, Stories \citep{trinh2018simple}, Books \citep{gao2020pile}, ArXiv \citep{gao2020pile}, and Code \citep{gao2020pile}. We consider models of different sizes: for \ssmname, this is indicated by the number of parameters in the model; for MegaByte, which is the primary baseline used, size is indicated by the number of parameters in the patched model and the unpatched generation head. Dataset sizes and average document lengths are included in Appendix~\ref{app:data}; model details are given in  Appendix~\ref{app:compute}.

Performance comparison across architectures requires care. To this end, we consider two settings: compute-matched and parameter-matched. This setup is necessary as the default MegaByte Transformer employs a global module that works with $8\times$-patched representations of the input, thus using $8\times$ fewer feed-forward FLOPs per byte than a raw Transformer, while having significantly more parameters. Table~\ref{tab:model-flops} shows the MegaByte and \ssmname model sizes employed in our experiments. The (forward pass) FLOPs computation for various model architectures and the associated hyperparameters employed are detailed in Appendix~\ref{app:compute}.


All \ssmname models were trained using the open-source Mamba code base.\footnote{\url{https://github.com/state-spaces/mamba}.} At training, we shuffle the documents and use contiguous sequences of $8,192$ bytes (one per document), starting from a random position. We enable mixed precision training using BF$16$ for training efficiency at scale. The optimizer, learning rate scheduler, and other training details are specified in Appendix~\ref{app:training}. 


\begin{wrapfigure}{r}{0.5\textwidth}
\vspace{-2em}
    \centering
    \includegraphics[width=\textwidth]{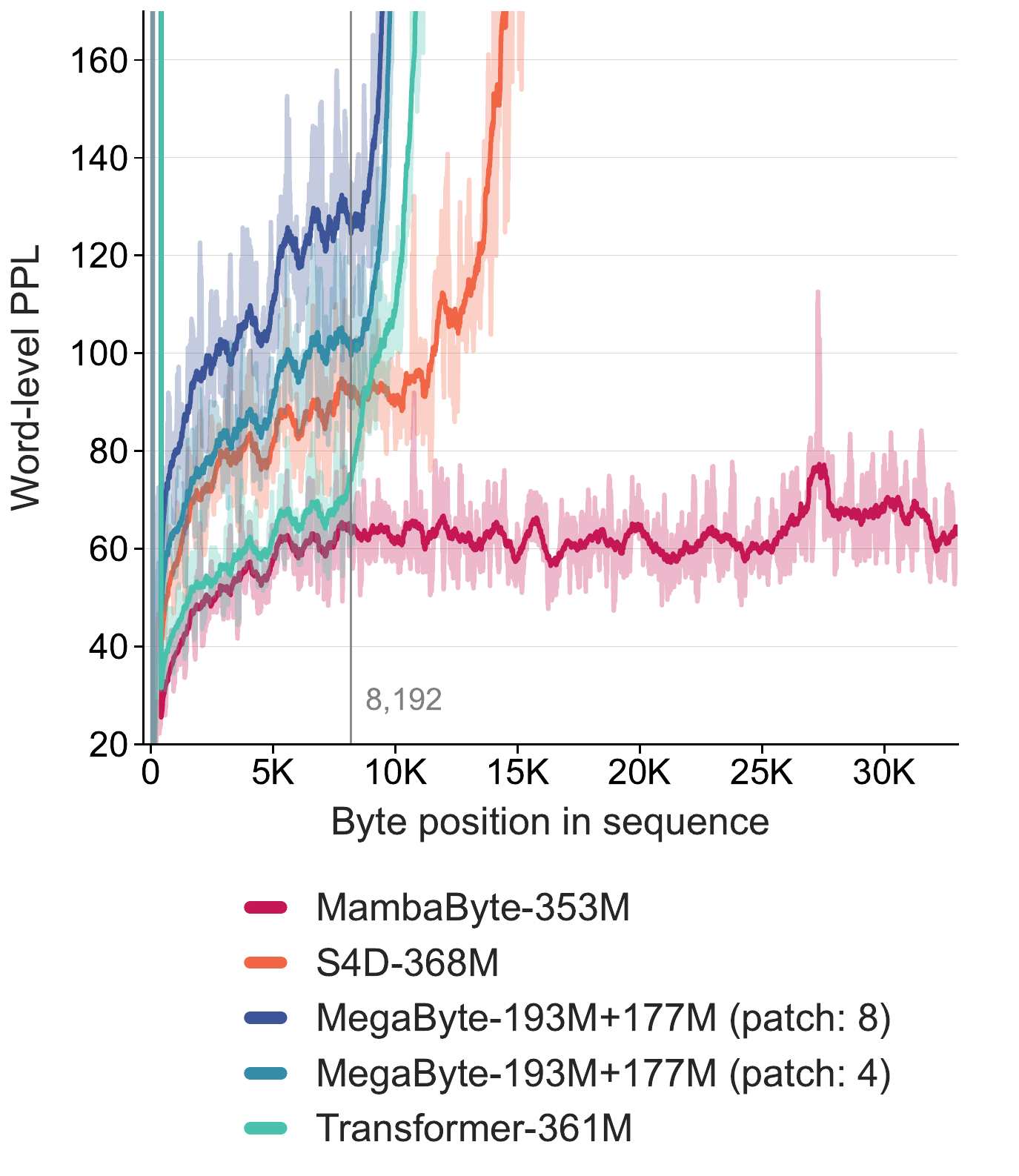}
    \caption{\textbf{Length extrapolation.} 
    All models are trained with $8,192$-long byte sequences. 
    \ssmname can extrapolate to much longer sequences without performance degradation.}
    \label{fig:len-extra}
\end{wrapfigure}

\section{Results}
\label{sec:results}

\subsection{Language modeling performance} 
\label{sec:6.1}

Table~\ref{tab:byte-results} shows language modeling performance in bits per byte ($\bpb$) across each dataset. For this experiment, the \megabytesmall and \ssmname models use the same number of FLOPs per byte (see Table~\ref{tab:model-flops}). We observe \ssmname to outperform MegaByte consistently across all datasets. Furthermore, \ssmname outperforms MegaByte with $0.63\times$ less compute and training data. Additionally, \smallssm also outperforms byte-level Transformer and PerceiverAR.

\begin{table}[!t]
    \centering
    \bgroup
    \def\arraystretch{1.07}
    \resizebox{\columnwidth}{!}{
    \begin{tabular}{l@{\hspace{1.1\tabcolsep}}ccccccc}
    \toprule
       \multirow{2}{*}{Byte-level model} & \multirow{2}{*}{Context} & \multirow{2}{*}{\makecell{Bytes\\trained}}~~ & \multicolumn{5}{c}{Test $\bpb\downarrow$ } \tabularnewline
       \cmidrule(lr){4-8}
        & & & PG19 & Stories & Books & ArXiv & Code \tabularnewline
    \midrule
       Transformer-$320$M &  $1,024$ & $80$B~~ &  $1.057$ & $1.064$ & $1.097$ & $0.816$ & $0.575$ \tabularnewline
       PerceiverAR-$248$M & $8,192$ & $80$B~~ & $1.104$ & $1.070$ & $1.104$ & $0.791$ & $0.546$ \tabularnewline
       \megabytesmall (patch: $8$) & $8,192$ & $80$B~~ & $1.000$ & $0.978$ & $1.007$ & $0.678$ & $0.411$ \tabularnewline
       \smallssm & $8,192$ & $30$B$^\ast$ & $\mathbf{0.930}$ & $\mathbf{0.908}$ & $\mathbf{0.966}$ & $\textbf{0.663}$ & $\mathbf{0.396}$ \tabularnewline
    \bottomrule
    \end{tabular}}
    \caption{\textbf{Medium-scale token-free experiments.} 
    \megabytesmall and \smallssm use the same FLOPs per byte. 
    (The $\bpb$ for Transformer, PerceiverAR, and MegaByte are from \citet{yu2023megabyte}.)}
    \label{tab:byte-results}
    \egroup
\end{table}




Figure~\ref{fig:pg19-test-performance} further explores this relationship by looking at models with the same number of parameters. The graphs indicate that for MegaByte models of the same parameter size, models with less input patching perform better, but when compute-normalized, they perform similarly. In fact, a full-length Transformer, while slow in an absolute sense, also performs similarly to the MegaByte model when compute-normalized. In contrast, switching to the Mamba architecture significantly improves both the compute usage and the model performance.

Following these findings, Table~\ref{tab:ppl-results} compares a larger version of these models on the PG19 dataset, both with and without tokenization. For this experiment, we compare \mediumssm with \megabytelarge and other byte-level models, as well as several state-of-the-art subword models. (The conversion from $\bpb$ and subword-level perplexity to word-level perplexity ($\ppl$) is detailed in Appendix~\ref{app:metrics}). We find that \mediumssm, even just trained for $150$B bytes, outperforms all the byte-level models and achieves competitive performance with subword models.

Figure~\ref{fig:len-extra} records another interesting aspect of \ssmname: its ability to extrapolate to significantly longer sequences (at least $4\times$ longer than the training length) compared to other byte-level Transformer and SSM baselines, suggesting that \ssmname can effectively refine the recurrent hidden state for significantly longer sequences. As expected, limited by the position embeddings, Transformer models don't extrapolate beyond the training length.

\begin{table}[!t]
    \centering
    \bgroup
    \def\arraystretch{1.07}
    \resizebox{\columnwidth}{!}{
    \begin{tabular}{c@{\hspace{0.7\tabcolsep}}|l@{\hspace{0.6\tabcolsep}}c@{\hspace{0.8\tabcolsep}}l@{\hspace{0.6\tabcolsep}}c@{\hspace{0\tabcolsep}}c@{\hspace{\tabcolsep}}c}
    \toprule
        \multicolumn{1}{c}{} & ($\#$Layers) Model & Vocab & \makecell[l]{Effective ctx\\(in bytes)} & 
        \makecell{Effective\\bytes\\trained}~~ &
        \makecell{Val\\$\ppl$}$\downarrow$ & \makecell{Test\\$\ppl$}$\downarrow$ \tabularnewline
    \midrule
        \multirowrotate{5}{90}{Subword} & ($36$) Transformer-XL \citep{rae2020compressive} & $32$K & $2,048/4,096$ & $400$B~~ & $45.5$ & $36.3$\tabularnewline
         & ($36$) Compressive \citep{rae2020compressive} & $32$K & $2,048/2$$\times 2,048$ & $400$B~~ & $43.4$ & $33.6$ \tabularnewline
         & ($22$) Routing-$490$M \citep{roy2021efficient} & $82$K & $32,768$ & $330$B~~ & $-$ & $33.2$ \tabularnewline
         & ($60$) PerceiverAR-$974.6$M \citep{hawthorne2022general} & $32$K & $8,192$ & $1.68$T~~ & $45.9$ & $28.9$ \tabularnewline
         & ($24$) Block-Recurrent-$1.3$B \citep{hutchins2022block} & $32$K & $4,096/$recurrence & $-$~~ & $-$ & $\mathbf{26.5}$ \tabularnewline
         & ($48$) Mamba-$1.03$B & $32$K & $8,192$ & $600$B$^\ast$ & $40.1$ & $33.9$  \tabularnewline
    \midrule
        \multirowrotate{4}{90}{Byte} & ($-$) Transformer-$320$M \citep{yu2023megabyte} & $256$ & $8,192$ & $400$B~~  & $81.6$ & $69.4$ \tabularnewline
         & ($-$) PerceiverAR-$248$M \citep{yu2023megabyte} & $256$ & $8,192$ & $400$B~~ & $119.1$ & $88.8$ \tabularnewline
         & ($24$+$24$) \megabytelarge \citep{yu2023megabyte} & $256$ & $8,192/$patch: $8$ & $400$B~~ & $42.8$ & $36.4$ \tabularnewline
         & ($\mediumssmlayers$) \mediumssm & $256$ & $8,192$ & $150$B$^\ast$ & $\mathbf{39.6}$ & $33.0$
        \tabularnewline
    \bottomrule
    \end{tabular}}
    \caption{\textbf{Large-scale experiment on PG19.} 
    The observed $\bpb$ scores are converted to word-level $\ppl$ for comparison with past works. All the byte-level models are compute-matched. Mamba-$1.03$B and \mediumssm are evaluated using $4\times$ longer context and a sliding window of $16,384$ bytes. \mediumssm significantly outperforms other byte-level models and is competitive with state-of-the-art subword models.
    (Accompanying citation indicates the work from which the corresponding result was taken; fields marked $-$ are unknown.)}
    \label{tab:ppl-results}
    \egroup
\end{table}

\subsection{Token-free capabilities}
\label{sec:6.2}

\begin{figure}
    \centering
    \begin{floatrow}
    \ffigbox{%
        \includegraphics[width=0.45\textwidth]{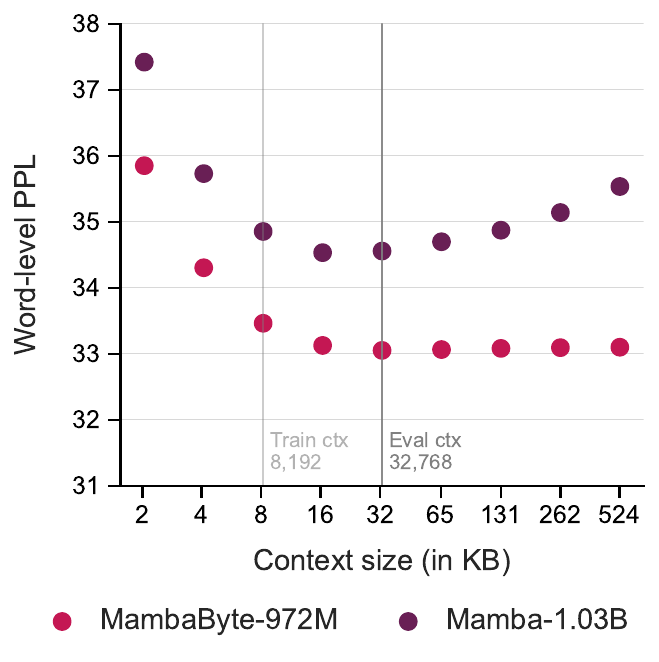}
    }{\caption{\textbf{Long context experiment.}
    Length extrapolation using a sliding window of $L_\text{ctx}/2$.}
    \label{fig:chunk-perf}}
    \capbtabbox{%
        \bgroup
        \def\arraystretch{1.07}
        \resizebox{0.45\textwidth}{!}{
        \begin{tabular}{lcc@{\hspace{0.5\tabcolsep}}c}
        \toprule
           \multirow{2}{*}{} & \multirow{2}{*}{Proba} & \multicolumn{2}{c}{$\ppl\downarrow$} \tabularnewline
           \cmidrule(lr){3-4}
            & & Mamba & MambaByte \tabularnewline
        \midrule
           \multirow[t]{4}{*}{Drop} & $0.05$ & $+16.9$ & $\mathbf{+8.5}$ \tabularnewline
            & $0.3$ & $+213.2$ & $\mathbf{+31.7}$ \tabularnewline
        \midrule
           \multirow[t]{4}{*}{Repeat} & $0.05$ & $+6.3$ & $\mathbf{+6.2}$ \tabularnewline
            & $0.3$ & $+28.4$ & $\mathbf{+26.6}$ \tabularnewline
        \midrule
           \multirow[t]{2}{*}{Antspeak} &  & $+58300.0$ & $\mathbf{+28.3}$ \tabularnewline
        \midrule
           \multirow[t]{2}{*}{Uppercase} & $0.05$ & $+5.4$ & $\mathbf{+1.6}$ \tabularnewline
            & $0.3$ & $+18.3$ & $\mathbf{+5.5}$ \tabularnewline
        \midrule
           \multirow[t]{2}{*}{Random case} & & $+20.8$ & $\mathbf{+7.7}$ \tabularnewline
        \midrule
           \multirow[t]{2}{*}{Swap} & $0.05$ & $+29.0$ & $\mathbf{+9.3}$ \tabularnewline
            & $0.3$ & $+630.6$ & $\mathbf{+28.7}$ \tabularnewline
        \bottomrule
        \end{tabular}}
        \egroup
        \vspace{1.2em}
    }{\caption{\textbf{Noise experiments.} 
    Degradation of Mamba-$1.03$B and \mediumssm under varied noise settings.}
    \label{tab:noise-results}}
    \end{floatrow}
\end{figure}

To control for the benefits of the Mamba architecture, we retrained a subword Mamba-$1.03$B model in a compute-matched setting (see Table~\ref{tab:ppl-results}). Interestingly, the (subword) Mamba and \ssmname perform similarly at the same parameter size and training compute. As previously mentioned, these models effectively have the same memory capacity despite significant differences in the input sequence length. We also find that Mamba achieves near-optimal performance more efficiently than MambaByte, though not $4\times$ faster as expected, but $2.2\times$ faster. Furthermore, the perplexity for Mamba-$1.03$B does not improve significantly beyond $150$B training bytes, consistent with the observations made by \citet{rae2020compressive}. Given the similar performance of Mamba and \ssmname, we can further explore downstream capabilities.





\paragraph{Modeling longer contexts.} From Figure~\ref{fig:chunk-perf}, we note that Mamba and \ssmname show impressive extrapolation capabilities for sequences up to $64\times$ longer than the training length. We hypothesize that \ssmname shows slightly better length extrapolation than the subword Mamba because \ssmname models $4\times$ longer sequences at training despite both models processing the same effective number of bytes per training sequence.

\paragraph{Synthetic noise experiments.} We employ the synthetic noise benchmark from \citet{xue2022byt5} to test model robustness; additional details about the noise settings are noted in Appendix~\ref{app:noise}. We process the input text in the PG19 test set into chunks of $100$ space-separated words and inject noise into every odd-indexed chunk while retaining the text in the even-indexed chunk unaltered. Table~\ref{tab:noise-results} shows the degradation of word-level $\ppl$ with noise injection, measured on even-indexed chunks. We observe that Mamba performance degrades significantly in the presence of noise compared to \ssmname across all noise conditions, indicating that tokenized vocabulary fundamentally limits subword models. This effect is pronounced in specific noise settings such as antspeak (every character is capitalized and padded with spaces) and character swapping (consecutive bytes are swapped). Our findings align with those observed by \citet{xue2022byt5} in that byte-level models are significantly more resilient to accidental and adversarial spelling errors than subword models.


\subsection{Generation efficiency}
\label{sec:6.3}


Autoregressive inference in Transformer models requires caching the entire context, which can significantly affect the generation speed. \ssmname does not suffer from this bottleneck as it maintains a single hidden state per layer that evolves with time, enabling constant time per generation step. Table~\ref{tab:speed-results} compares the text generation speeds of \mediumssm and \bigssm with \megabytelarge on an A100 80GB PCIe GPU. While MegaByte significantly reduces the generation cost through patching, we observe \ssmname to be $2.6\times$ faster in a parameter-matched setting due to its use of recurrent generation. Appendix~\ref{app:samples} includes more information about the generation process.



\begin{table}[!t]
    \centering
    \bgroup
    \def\arraystretch{1.07}
    \resizebox{0.8\columnwidth}{!}{
    \begin{tabular}{lcccc}
    \toprule
       Model & \makecell{Bytes\\trained}~~ & Context & \makecell{Test\\$\bpb$} $\downarrow$ & \makecell{Generation\\time (s)} $\downarrow$ \tabularnewline
    \midrule
        Transformer-$350$M & $80$B & $1,024$ & $1.064$ & $132$ \tabularnewline
        \megabytemedium  & $80$B & $8,192$ & $0.991$ & $93$ \tabularnewline
    \tablerule
        \megabytemedium & $-$~~ & $8,192$ & $-$ & 265 \tabularnewline
    
       \mediumssm & \multirow[t]{2}{*}{~$75$B$^\ast$} & $8,192$ & $\mathbf{0.883}$ & $\mathbf{29}$ \tabularnewline
       ~~~w/ sliding window ($2\times$ bytes) & & & $\mathbf{0.863}$ & 58 \tabularnewline
       \bigssm & $-$~ & $8,192$ & $-$ & $36$
       \tabularnewline
    \bottomrule
    \end{tabular}}
    \caption{\textbf{Generation speed benchmarking.} 
    Speed to generate $8,192$ bytes; fields marked $-$ are unknown. 
    (Upper) The $\bpb$ on PG19 and generation time for the Transformer and MegaByte are taken from \citet{yu2023megabyte}.
    (Lower) MegaByte and MambaByte run on the same hardware; we use the available open-source MegaByte implementation here.}
    \label{tab:speed-results}
    \egroup
\end{table}

\paragraph{Generation via subword speculation.} Table~\ref{tab:spec-dec} shows the inference speedup using speculative decoding through subword drafting, averaged across $100$ prompts generated using common phrases in the PG19 dataset. We use a Mamba-$110$M model as the drafter, and the subwords are generated using greedy decoding. We observe that through speculative subword drafting, \ssmname can achieve a decoding speed nearing that of the subword Mamba. Furthermore, to assess the faithfulness of our speculative decoding approach, we use a greedy-decoded \mediumssm generation as the reference candidate for a given prompt. We report the ratio of the log-likelihood of generating the reference candidate to the log-likelihood of \ssmname generating the speculative-decoded sequence, which is averaged across all prompts. From Table~\ref{tab:spec-dec}, we observe our speculative decoding approach to be more faithful to \mediumssm than the subword Mamba-$1.03$B. 

\begin{table} 
    \centering
    \bgroup
    \def\arraystretch{1.07}
    \resizebox{0.7\columnwidth}{!}{
    \begin{tabular}{lccccc}
    \toprule
       Model & Context & \makecell{Relative\\speedup} $\uparrow$  & \makecell{Log-odds\\ratio} $\uparrow$ \tabularnewline
    \midrule
        \mediumssm & $8,192$ & $1\times$ & $1.0$ \tabularnewline
        Mamba-$1.03$B & $2,048$ & $2.8\times$ & $0.10$ \tabularnewline
    \tablerule
       \makecell[l]{\mediumssm\\~~~w/ Mamba-$110$M speculation}  & $8,192$ & $\mathbf{2.6\times}$ & $\mathbf{0.89}$ \tabularnewline
    \bottomrule
    \end{tabular}}
    \caption{\textbf{Generation speed with subword speculation.} 
    Empirical results for speeding up inference from the \mediumssm model. Speed was measured in generating $8,192$ bytes; the drafter drafts three subwords per iteration and the verifier accepts if the drafted bytes were in its top-$3$ candidates.}
    \label{tab:spec-dec}
    \egroup
\end{table}
\section{Related work}
\label{sec:rel}


\paragraph{Token-free language models.} Tokenization has been fundamental to language modeling and vital in enhancing model efficiency and understanding. Several algorithms have been proposed to address tokenization issues, including sizeable vocabulary size and handling out-of-vocabulary tokens: Byte-Pair Encoding \citep{sennrich2015neural}, WordPiece \citep{schuster2012japanese,devlin2018bert}, and SentencePiece \citep{kudo2018sentencepiece}. The recent shift towards character
\citep{tay2022charformer,ma2020charbert,mielke2019spell} and byte-level \citep{yu2023megabyte,xue2022byt5,belouadi2022bygpt5} modeling aims to achieve token-free preprocessing, thereby facilitating improved model adaptability and domain transferability in language modeling and multilingual processing.

\paragraph{Attention-free models.} Attention-free models offer enhanced computational and memory efficiency and are increasingly adapted for several language processing tasks, including autoregressive language modeling. Models such as S4 \citep{gu2021efficiently} and its subsequent variants \citep{gupta2022diagonal,gu2022parameterization} have demonstrated promising outcomes in subword-level language modeling. Gated SSM architectures such as GSS \citep{mehta2023long} and BiGS \cite{wang2022pretraining} incorporated a gating mechanism into SSMs for (bidirectional) language modeling. The recently introduced Mamba model \citep{gu2023mamba} posits that the unchanging dynamics of these methods fail to incorporate input-specific context selection within the hidden state, which might be crucial for tasks like language modeling. Mamba has been shown to outperform Transformers across model sizes and at scale. Alternatively, several other sub-quadratic model architectures \citep{yang2023gated,de2024griffin,arora2023zoology,arora2024simple,fu2024monarch} have also been proposed. Beyond language modeling, SSMs and Mamba have been applied in other modalities, including images \citep{yan2023diffusion}, audio \citep{goel2022s}, and bioinformatics \citep{schiff2024caduceus}.

\paragraph{Speculative decoding for fast inference.} Speculative decoding \citep{spector2023accelerating,leviathan2023fast,chen2023accelerating,xia2023speculative} has emerged as a promising approach to accelerate the inference of large language models, specifically Transformers. The core idea is to use a smaller draft model to speculatively generate candidate tokens, which the larger target model then verifies. \citet{leviathan2023fast,chen2023accelerating} proposed a rejection sampling scheme to improve the inference quality. \citet{spector2023accelerating} restructured the candidate tokens into a tree to enable more efficient verification. Additional approaches also investigated trained draft models \citep{bhendawade2024speculative,chen2023cascade,liu2023online} and training-free draft models \citep{he2023rest,yang2023inference,fu2024break}. While previous methods employ drafter and verifier models with the same underlying tokenization scheme, this paper proposes using a smaller subword Mamba model as the speculative drafter and a larger \ssmname as the byte-level verifier.
\section{Conclusion}
\label{sec:concl}

We introduce \ssmname, a token-free SSM for modeling long byte-sequences. \ssmname outperforms other byte-level models over several datasets and shows competitive results with subword Transformers while being significantly robust to text corruptions, thus serving as a promising tokenization alternative. Due to their recurrent nature, SSMs enable significantly faster text generation to Transformer models. We further improve the generation efficiency through speculative decoding using subword drafting and show \ssmname to achieve to achieve a decoding efficiency similar to the subword Mamba, making byte models practical. Our findings establish the possibility of token-free language modeling in future large models.

\ifauthorinfo%
    \section*{Acknowledgments}
\label{sec:ack}

We thank Albert Gu for their helpful comments on \ssmname, Tri Dao for their guidelines on extending the selective scan kernel in Mamba, and the authors of MegaByte, Lili Yu and Mike Lewis, for clarifications on MegaByte training and inference procedures. This work was supported by NSF IIS-$1901030$ and NSF CAREER $2037519$.
\fi

\bibliographystyle{conf/colm2024_conference}
\bibliography{references}

\newpage\onecolumn
\addtocontents{toc}{\protect\setcounter{tocdepth}{1}}
\tableofcontents\newpage
\begin{appendices}
    \section{Dataset specifics}
\label{app:data}

\begin{wraptable}{l}{0.5\textwidth}
\vspace{-1.3em}
    \centering
    \bgroup
    \def\arraystretch{1.07}
    \resizebox{\textwidth}{!}{
    \begin{tabular}{lccc}
    \toprule
        & Total bytes & Total docs & Bytes$/$doc \tabularnewline
    \midrule
       PG19 & $11.74$G & $28,752$ & $4,082,210$ \tabularnewline
       Stories & $34.18$G & $948,247$ & $36,045$ \tabularnewline
       Books & $108.38$G & $196,640$ & $551,179$ \tabularnewline
       ArXiv & $60.27$G & $1,264,405$ & $47,665$ \tabularnewline
       Code & $677$G & $56,626,342$ & $11,958$ \tabularnewline
    \bottomrule
    \end{tabular}}
    \caption{\textbf{Text dataset statistics.}
    The total bytes, total documents, and the mean document size (bytes per document) for each dataset.}
    \label{tab:data-stats}
    \egroup
\vspace{0.5em}
\end{wraptable}

We benchmark our results on various long-form text datasets. The PG19 dataset \citep{rae2020compressive} is an extensive collection of full-length English books (written before $1919$) from the Project Gutenberg online library. The PG19 dataset is ideal to test for long-distance context modeling \citep{gao2020pile}. The Stories dataset \citep{trinh2018simple} is a subset of the CommonCrawl data used for commonsense reasoning and language modeling. The Books dataset \citep{gao2020pile} is another collection of English books. The ArXiv dataset \citep{gao2020pile} comprises technical publications in \LaTeX\ from the arXiv online archive. Finally, the Code dataset \citep{gao2020pile} is a large dataset of publicly available open-source code (under Apache, MIT, or BSD licenses). Dataset statistics are tabulated in Table~\ref{tab:data-stats}.

For the PG19 dataset, we employ the train, validation, and test data splits as indicated by \citet{rae2020compressive}. For Stories, Books, ArXiv, and Code datasets, we randomly sample $40$M consecutive bytes for testing and the rest to train \ssmname.

    \section{Compute-constrained modeling}
\label{app:compute}


\begin{figure*}[!ht]
\centering
\resizebox{\textwidth}{!}{%
    \begin{subfigure}{0.8\columnwidth}
    \centering
        \includegraphics[]{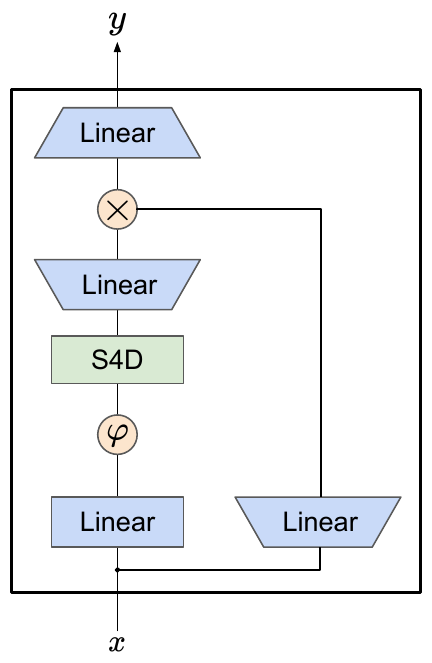}
    \label{fig:gated-s4d-arch}
    \end{subfigure}
\vline
    \begin{subfigure}{0.8\columnwidth}
    \centering
        \includegraphics[]{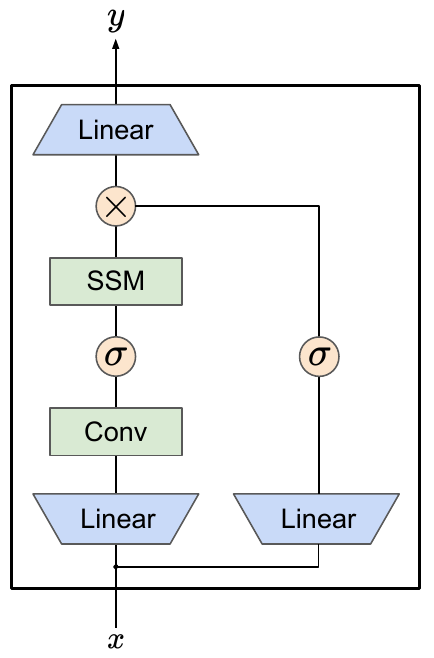}
    \label{fig:mamba-arch}
    \end{subfigure}}
\caption{\textbf{SSM model network architectures.} (Left) Gated-S4D block adapted from \citet{mehta2023long}; (right) Mamba SSM block. ($\varphi$ indicates GELU activation \citep{hendrycks2016gaussian}, and $\sigma$ indicates Swish activation \citep{ramachandran2017searching}.)}
\label{fig:model-arch}
\end{figure*}

As noted earlier, we evaluate and benchmark \ssmname in a compute-controlled setting. To this end, we estimate the FLOPs per byte incurred by various byte-level model architectures. We parameterize the architectures using hyperparameters $n$ $(n_g/n_l)$ number of (global$/$local) layers, dimension $d$ $(d_g/d_l)$ of the (global$/$local) residual stream, expansion factor $e$ of linear layers, patch size $p$ in MegaByte, state dimension $n_\text{state}$ in SSMs, 1D convolution kernel size $k$, and low-rank projection dimension $r$ in Mamba. We also include $L_\text{ctx}$ bytes in the input context. Detailed component-wise compute counts for the forward pass are included in Table~\ref{tab:flops}.

\begin{table}[!ht]
    \centering
    \bgroup
    \def\arraystretch{1.07}
    \resizebox{\columnwidth}{!}{
    \begin{tabular}{lll}
    \toprule
       Model & Component & FLOPs per byte \tabularnewline
    \midrule
        \multirow{2}{*}{\makecell[l]{Transformer\\\citep{vaswani2017attention}}} & Multi-head attention & $2n(4d^2 + 2L_\text{ctx}d)$ \tabularnewline
         & Pointwise feed-forward & $2n(2ed^2)$ \tabularnewline
    \tablerule
        \multirow{4}{*}{\makecell[l]{MegaByte$^\text{\ref{mb-oss}}$\\\citep{yu2023megabyte}}} & Embedding projection & $2d_g^2$ \tabularnewline
         & Global transformer model & $2n_g(4d_g^2 + 2d_gL_\text{ctx}/p + 2ed_g^2)/p$ \tabularnewline
         & Global-to-local projection & $2d_gd_l$ \tabularnewline
         & Local transformer model & $2n_l(4d_l^2 + 2pd_l + 2ed_l^2)$ \tabularnewline
    \tablerule
        \multirow{3}{*}{\shortstack[l]{Gated-S4D\\(Figure~\ref{fig:model-arch}, left)}} & Linear projections & $2n(3ed^2 + d^2)$ \tabularnewline
         & Kernel via Vandermonde $\varv(\discrete{\A})$ & $n(\alpha_\varv ed(n_\text{state} + L_\text{ctx})\log_2^2(n_\text{state} + L_\text{ctx})/L_\text{ctx})$ \tabularnewline
         & S4D SSM with convolution & $n(\alpha_\text{fft}\log(L_\text{ctx})ed + ed)$ \tabularnewline
         & Element-wise gating & $ned$ \tabularnewline
    \tablerule
        \multirow{8}{*}{\shortstack[l]{\ssmname\\(Figure~\ref{fig:model-arch}, right)}} & Linear projections & $2n(3ed^2)$ \tabularnewline
         & Pre-SSM 1D convolution & $2nked$ \tabularnewline
         & $\Delta, \B, \C$ from input $x$ & $2n (2edr + 2edn_\text{state})$ \tabularnewline
         & Discretization, pre-scan: $\discrete{\A}$, $\discrete{\B}x$ & $n(3edn_\text{state})$ \tabularnewline
         & Recurrence with parallel scan & $n(edn_\text{state})$ \tabularnewline
         & Output: $y = \discrete{\C}h + \discrete{\D}x$ & $2nedn_\text{state} + ned$ \tabularnewline
         & Element-wise gating & $ned$ \tabularnewline
    \bottomrule
    \end{tabular}}
    \caption{\textbf{Compute (forward pass) estimates for various byte-level language models.}
    Embedding, de-embedding, and sub-leading terms such as biases, nonlinearities, and layer norms are omitted. ($\alpha_\ast$ indicates an implementation-specific constant scaling term.)}
    \label{tab:flops}
    \egroup
\end{table}

\begin{figure}[!ht]
    \centering
    \includegraphics[width=\textwidth]{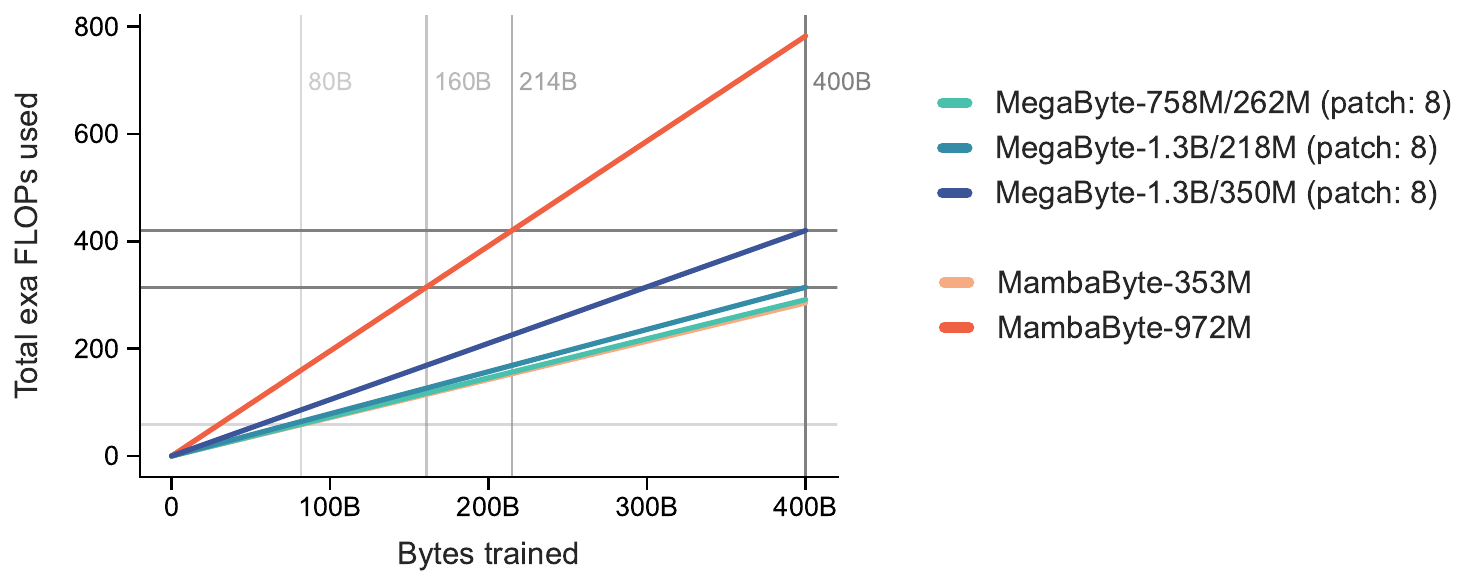}
    \caption{\textbf{Computational cost for different model architectures at different scales.} All models use a context length of $8,192$, and MegaByte architectures use a patch size of $8$.}
    \label{fig:flops-comparison}
\end{figure}

For the medium-scale language modeling experiments (Table~1, $\S5$ of \citet{yu2023megabyte}), \citet{yu2023megabyte} employ the \megabytesmall model, with a context length of $8,192$ and patch size of $8$, trained for $80$B bytes. As shown in Figure~\ref{fig:flops-comparison}, \smallssm (\smallssmconfig) and \megabytesmall use the same total compute in FLOPs; hence, we employ the \smallssm to benchmark against \megabytesmall in Table~\ref{tab:byte-results} of $\S\ref{sec:results}$.

For the PG19 scaling experiment (Table~2, $\S5$ and Appendix~$\mathrm{D}.3$ of \citet{yu2023megabyte}), \citet{yu2023megabyte} use \megabytelarge (context length of $8,192$ and patch size of $8$) trained for $400$B bytes to benchmark the observed word-level perplexity against several state-of-the-art subword models. Owing to our hardware limitations, we train \mediumssm (\mediumssmconfig) and control for the total compute used (see Figure~\ref{fig:flops-comparison} to view the associated computational costs). All the model sizes and associated hyperparameters employed in this work are tabulated in Table~\ref{tab:hyperparams}.

\begin{table}[!ht]
    \centering
    \bgroup
    \def\arraystretch{1.07}
    \resizebox{\columnwidth}{!}{
    \begin{tabular}{l@{\hspace{\tabcolsep}}l@{\hspace{0.6\tabcolsep}}c@{\hspace{1.1\tabcolsep}}c@{\hspace{1.1\tabcolsep}}c@{\hspace{1.1\tabcolsep}}c@{\hspace{1.3\tabcolsep}}l}
    \toprule
       \multirow{2}{*}{Model} & \multirow{2}{*}{Parameters} & \multicolumn{5}{c}{Hyperparameters} \tabularnewline
       \cmidrule(lr){3-7}
        & & \makecell{$n$\\$(n_g/n_l)$} & \makecell{$d$\\$(d_g/d_l)$} & $e$ & $L_\text{ctx}$ & Others \tabularnewline
    \midrule
        \multirow{2}{*}{Transformer} & $320$M \citep{yu2023megabyte} & $22$ & $1,024$ & $4$ & $1,024$ & heads: $-$ \tabularnewline
         & $350$M \citep{yu2023megabyte} & $24$ & $1,024$ & $4$ & $1,024$ & heads: $16$  \tabularnewline
         & $361$M & $28$ & $1,024$ & $4$ & $8,192$ & heads: $16$ \tabularnewline
    \tablerule
        PerceiverAR & $248$M \citep{yu2023megabyte} & $17$ & $1,024$ & $4$ & $8,192$ & latents: $1,024$   \tabularnewline
    \tablerule
        \multirow{2}{*}{MegaByte} & $193$M+$177$M\tablefootnote{\label{mb-oss}We used the open-source implementation: \url{https://github.com/lucidrains/MEGABYTE-pytorch}.} & $14/14$ & $1,024/1,024$ & $4$ & $8,192$ & $p = 4, 8$; heads: $16/16$ \tabularnewline
         & $758$M+$262$M \citep{yu2023megabyte} & $14/18$ & $2,048/1,024$ & $4$ & $8,192$ & $p = 8$; heads: $16/16$ \tabularnewline
         & $1.3$B+$218$M \citep{yu2023megabyte} & $24/15$ & $2,048/1,024$ & $4$ & $8,192$ & $p = 8$; heads: $32/-$ \tabularnewline
         & $1.3$B+$350$M \citep{yu2023megabyte} & $24/24$ & $2,048/1,024$ & $4$ & $8,192$ & $p = 8$; heads: $32/16$ \tabularnewline
    \tablerule
        Gated-S4D & $368$M & $26$ & $1,024$ & $4$ & $8,192$ & $n_\text{state} = 64$ \tabularnewline
    \tablerule
        \multirow{2}{*}{\ssmname} & \smallssmsize & \smallssmlayers & \smallssmdim & \smallssmexp & $8,192$ & $k = 4; n_\text{state} = 16; r = 64$ \tabularnewline
         & \mediumssmsize & \mediumssmlayers & \mediumssmdim & \mediumssmexp & $8,192$ & $k = 4; n_\text{state} = 16; r = 112$ \tabularnewline
         & \bigssmsize & \bigssmlayers & \bigssmdim & \bigssmexp & $8,192$ & $k = 4; n_\text{state} = 16; r = 144$ \tabularnewline
    \bottomrule
    \end{tabular}}
    \caption{\textbf{Model hyperparameters.}
    We report the model size and associated hyperparameters for all the models employed in this study. (Accompanying citation indicates the work from which the associated configuration is noted; fields marked as $-$ are unknown.)}
    \label{tab:hyperparams}
    \egroup
\end{table}

    \section{Training recipes}
\label{app:training}


All the models in this study were trained using an AdamW optimizer with $\beta = (0.9, 0.95)$. We used a linear learning rate warm-up (for the first $500$ steps) followed by cosine annealing. Keeping consistent with MegaByte training \citep{yu2023megabyte}, we used a batch size of $48$ across all our experiments. Additionally, we do not use dropout with any of our models.

For the experiments in Figure~\ref{fig:pg19-test-performance}, we conducted a hyperparameter search using peak learning rates of $0.0002$, $0.0006$, and $0.0008$ and clipped the gradient norm to $1.0$ for all the models. The best-observed performance curve for each model is reported in Figure~\ref{fig:pg19-test-performance}. Furthermore, we use an improved Transformer recipe that uses RMSNorm instead of LayerNorm, rotary positional encodings \citep{su2021roformer}, and linear terms without bias (same as \citet{yu2023megabyte}).




In our medium-scale experiments shown in Table~\ref{tab:byte-results}, we set the peak learning rate to $0.0004$ and clipped the gradient norm to $0.1$. We trained the \smallssm for a total of $80$K steps, equivalent to $80,000 \times 48 \times 8,192 \approx 30$B bytes.

In the large-scale experiment on PG19, we use a similar setting to that in the medium-scale experiments: the peak learning rate is set to $0.0004$, and the gradient norm is clipped to $0.1$. The \mediumssm is trained for $380$K steps, equivalent to $380,000 \times 48 \times 8,192 \approx 150$B bytes. 




    \section{Discretization and selection}
\label{app:selection}

\begin{figure}[!t]
    \centering
    \includegraphics[width=\textwidth]{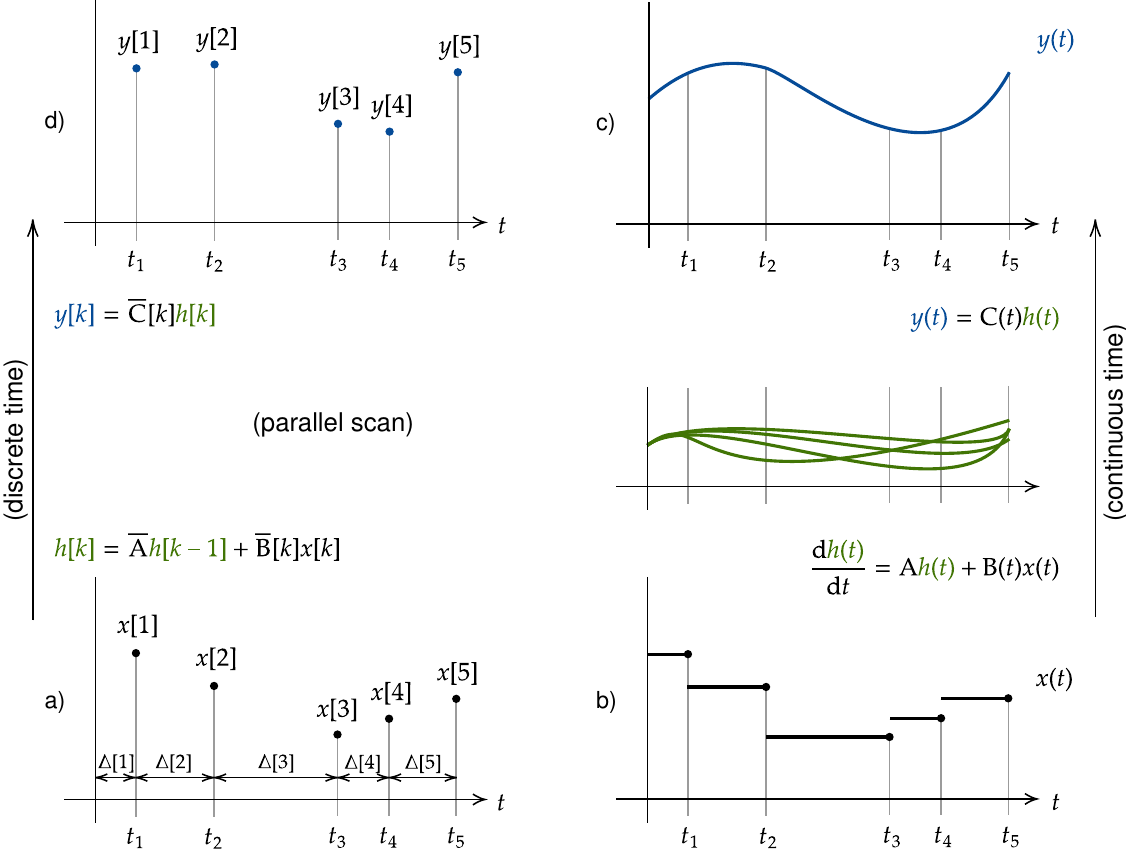}
    \caption{\textbf{Illustration of the Mamba SSM.} 
    (a) The discrete-time input $x[k]$, along with input-selective $\Delta[k]$.
    (b) The continuous-time signal $x(t)$.
    (c) Mathematically, the SSM transforms the continuous-time $x(t)$ through an $n$-dimensional hidden state (here, $n = 4$) using parameters $\A$ and $\B(t)$, which is then mapped to the output $y(t)$ using $\C(t)$. 
    (d) Practically, we compute $y[k]$ using a discrete-time parallel scan at the steps defined by $\Delta[k]$ and discrete-time matrices $\discrete{\A}[k]$, $\discrete{\B}[k]$, and $\discrete{\C}[k]$. At inference, we run the recurrence directly.}
    \label{fig:mamba-explanation}
\end{figure}


Discretization has deep connections to continuous-time systems, which allows for desirable properties such as model normalization \citep{orvieto2023resurrecting,gu2023how} and resolution invariance \citep{nguyen2022s4nd}. In this section, we review how zero-order hold discretization of Mamba selective SSM can be viewed as a generalization of the gating mechanism in recurrent networks. An illustration of the Mamba SSM discretization and illustration is depicted in Figure~\ref{fig:mamba-explanation}.

\paragraph{Zero-order hold discretization.} For a given input $x(t) \in \R$, we wish to discretize a continuous-time SSM defined by (\ref{eq:mamba}) in $\S\ref{sec:model}$. To this end, we sample the system at different time intervals such that $x[k] = x(t_k)$ for $t_k = \sum_{j=1}^k \Delta[j]$ and assume a zero-order hold, i.e., $x(t)$ is constant between samples: $x(t_k + \xi) = x(t_k) = x[k]$ for any $\xi \in [t_k, t_{k+1})$. The resultant matrices of the associated discrete SSM are:\footnote{In Mamba \citep{gu2023mamba}, $\B$ is discretized through a simplified Euler (as opposed to zero-order hold) discretization from empirical observations of $\A$ being more important than $\B$, and the performance does not change significantly with simplification on $\B$.}
\begin{align*}
    \discrete{\A} = \exp(\A\Delta); \quad \discrete{\B} = \A^{-1} (\exp(\A\Delta) - \I) \B; \quad \discrete{\C} = \C.
\end{align*}

\paragraph{Selection mechanics and gating in recurrent networks.} \citet{gu2023mamba} note that a selective SSM can be realized as a gated recurrence by setting $\Delta = \softplus(z(x)) = \softplus(\linear{\Delta}{(\linear{R}{x})})$ (as indicated in (\ref{eq:selective-ssm}) of $\S\ref{sec:model}$). By letting $\A = -1$, $\B = 1$, and $n = 1$, the authors observe:
\begin{multicols}{2}
\noindent
    \begin{align*}
        \discrete{\A} &= \exp(\A\Delta) \\
        &= \exp(-\log(1 + \exp(z(x))))\\
        &= \frac{1}{1 + \exp(z(x))} \\
        &= \sigma(-z(x)) \\
        &= 1 - \sigma(z(x)).
    \end{align*}
    \begin{align*}
        \discrete{\B} &= \A^{-1} (\exp(\A\Delta) - \I) \B \\
        &= \I - \exp(\A\Delta) \\
        &= \sigma(z(x)).
    \end{align*}
\end{multicols}
\vspace{-0.9em}

Using $\discrete{\A}$ and $\discrete{\B}$ from above in the discrete recurrence (\ref{eq:discrete-ssm}), the selective SSM takes the form of a 1D gated recurrence:
\begin{align}
    \label{eq:selection-gating}
    h[k] &= \left(1 - \sigma(z(x))\right) h[k - 1] + \sigma(z(x)) x[k].
\end{align}

It is interesting to note from (\ref{eq:selection-gating}) that $\lim_{\Delta \to \infty} h[k] = x[k]$ and $\lim_{\Delta \to 0} h[k] = h[k - 1]$: a large $\Delta$ ($\Delta \to \infty$) denotes the evolution of the system to focus only on the current input and forgetting the state. In contrast, a small $\Delta$ ($\Delta \to 0$) represents a transient input being ignored.

\paragraph{Selectivity of $\boldsymbol{\A}$, $\boldsymbol{\B}$, and $\boldsymbol{\C}$ matrices.} \citet{gu2023mamba} argue that since the system matrix $\A$ only affects the model through $\Delta$, i.e., $\discrete{\A} = \exp(\A\Delta)$. Hence, the selectivity in $\Delta$ is sufficient to ensure selectivity in $\A$.

While the selectivity in $\Delta$ enables selectivity in the input matrix $\B$, \citet{gu2023mamba} hypothesize that making $\B$ and $\C$ selective (in addition to $\Delta$) would allow for more fine-grained control based on the content $x[k]$ and evolving context $h[k]$.

    \section{Evaluation metrics}
\label{app:metrics}

Subword-based language models \citep{vaswani2017attention,hawthorne2022general,hutchins2022block} report their performance in word or subword-level $\ppl$, while byte-level language models \citep{xue2022byt5,yu2023megabyte} report theirs in $\bpb$. To facilitate meaningful comparisons, we report performance in $\bpb$ when benchmarking against byte-level models and word-level $\ppl$ when comparing to token-level models.\footnote{Unless stated otherwise, we use $\ppl$ to report word-level (not subword-level) perplexity.} This section details the conversion from $\bpb$ and subword-level $\ppl$ to word-level $\ppl$.

Irrespective of the underlying segmentation, the amount of information $I(D)$ in a given dataset $D$ is constant. Simply put,
\begin{subequations}
\label{eq:bpb-conversion}
\begin{align}
    I(D) &= L_W\,\text{bits per word} = L_S\,\text{bits per subword} = L_B\,\text{bits per byte} \\
    &\triangleq \frac{-\ln(D; \text{model})}{\ln(2)},
\end{align}
\end{subequations}
where $L_W$, $L_S$, and $L_B$ are the length of the dataset in words, subwords, and bytes, respectively. From (\ref{eq:bpb-conversion}), we observe:
\begin{align*}
    \bpb = \frac{-\ln(D; \text{model}) / L_B}{\ln(2)} = \frac{\ell_\text{byte}}{\ln(2)},
\end{align*}
where $\ell_\text{byte}$ is the observed byte-level negative log-likelihood loss (computed using $\ln$). From (\ref{eq:bpb-conversion}), we also note the following conversion from $\bpb$ to word-level $\ppl$:
\begin{align*}
    \frac{-\ln(D; \text{model}) / L_W}{\ln(2)} &= \frac{L_B}{L_W} \bpb = \frac{L_B}{L_W} \frac{\ell_\text{byte}}{\ln(2)} \\
    \Rightarrow \ppl &= \exp\left(\frac{L_B}{L_W} \ell_\text{byte}\right) = \exp\left(\frac{L_B}{L_W} \ln(2) \bpb\right).
\end{align*}

Similarly, we can compute word-level $\ppl$ from the observed subword-level negative log-likelihood loss $\ell_\text{subword}$ as:
\begin{align*}
    \ppl = \exp\left(\frac{L_S}{L_W}\ell_\text{subword}\right).
\end{align*}

\begin{table}[!t]
    \centering
    \bgroup
    \def\arraystretch{1.07}
    \resizebox{0.9\columnwidth}{!}{
    \begin{tabular}{lccccc}
    \toprule
        & $L_B$ & $L_S$ & $L_W$ & $L_B/L_W$ & $L_S/L_W$ \tabularnewline
    \midrule
       Train & $11,677,824,216$ & $2,914,582,573$ & $1,973,048,393$ & $5.92$ & $1.48$ \tabularnewline
       Validation & $17,733,002$ & $4,357,506$ & $3,007,061$ & $5.90$ & $1.45$ \tabularnewline
       Test & $41,289,101$ & $10,282,006$ & $6,965,511$ & $5.93$ & $1.48$ \tabularnewline
    \bottomrule
    \end{tabular}}
    \caption{\textbf{PG19 dataset statistics.}
    Split-wise UTF-$8$ encoded byte $L_B$, SentencePiece-tokenized subword $L_S$, and space-separated word $L_W$ counts in the PG19 dataset. (The byte count includes the newline character.) We also indicate the associated bytes per word $L_B/L_W$ and subwords per word $L_S/L_W$.}
    \label{tab:pg19-lt-lb}
    \egroup
\end{table}

For the PG19 dataset, we train \mediumssm to minimize BPB over the training data and report word-level PPL on the test data. In our medium-scale benchmarking experiments, for (subword) Mamba-$1.03$B, we trained a $32$K-subword vocabulary using the SubwordTextEncoder from the \texttt{tfds} package in TensorFlow, the same as \citet{rae2020compressive}. Split-wise values of $L_B/L_W$ and $L_S/L_W$ for the PG19 dataset are tabulated in Table~\ref{tab:pg19-lt-lb}.

    \section{Speculative decoding through subword drafting algorithm}
\label{app:spec}

Algorithm~\ref{alg:spec} below outlines our speculative decoding approach: the smaller subword draft model drafts $m$ subwords at a time, which are then verified at a byte-level by a larger MambaByte model. 


\begin{algorithm}[!ht]
    \setstretch{1}
    \begin{algorithmic}
        \State \textbf{Inputs:} $M_\text{subword}$, $M_\text{byte}$, prefix subwords $s_{1:t}$, previous $M_\text{subword}$ hidden state $\hbar_\text{prev}$, previous $M_\text{byte}$ hidden state $h_\text{prev}$, draft block size $m$, verify model tolerance $n$.
        
        \LComment{Sample $m$ draft subwords $\tilde{s}_j$s from $M_\text{subword}$ autoregressively and record the hidden states $\hbar_j$s at each timestep.}
        \State $\hbar_0 \gets \hbar_\text{prev}$
        \For{$j = 1, \dotsc, m$}
            \State $q_j(x), \hbar_j \gets M_\text{subword}(s_{1:t} + \tilde{s}_{1:j-1}, \hbar_{j-1})$
            \State $\tilde{s}_j \sim q_j(x)$
        \EndFor

        \State $b_{1:t'}, \tilde{b}_{1:n} \gets \text{bytes}(s_{1:t}), \text{bytes}(\tilde{s}_{1:m})$ \Comment{Get bytes for both prefix and drafted subwords.}
        
        \LComment{Run $M_\text{byte}$ in parallel to verify the drafted bytes, while recording the associated hidden states $h_i$s.}
        \State $\left(p_1(x), h_1\right), \dotsc, \left(p_n(x), h_n\right) \gets\\
        ~~~~~~~~M_\text{byte}(b_{1:t'}, h_\text{prev}), M_\text{byte}(b_{1:t'} + \tilde{b}_1), \dotsc, M_\text{byte}(b_{1:t'} + \tilde{b}_{1:n-1})$
        \LComment{Find the bifurcation position $c$ such that $\tilde{b}_{1:c}$ drafted bytes all fall in top-$\beta$ candidates of $M_\text{byte}$, while $\tilde{b}_{c+1}$ does not.}
        \State $c \gets \min\left(\{i \,\vert\, 1 \leq i \leq n,\, \text{rank}_{p_i}(\tilde{b}_i) > \beta\} \cup \{n\}\right)$
        \State $c' \gets \min\left(\{j \,\vert\, 1 \leq j \leq m,\, \text{cumsum}(\text{len}(\tilde{s}_{1:m}))[j] > c\} \cup \{m\}\right)$ \Comment{Find associated subword bifurcation position.}
        
        \LComment{Starting from $\tilde{b}_c$ (and using $h_c$), sample corrected bytes $\hat{b}_i$s from $M_\text{byte}$ autoregressively until a boundary byte (e.g., space) is obtained.}
        \State $\hat{b}_c \gets \tilde{b}_c; k \gets 0$
        \While{$\hat{b}_{c+k}$ is not a boundary byte}
            \State $k \gets k + 1$
            \State $p_{c+k}(x), h_{c+k} \gets M_\text{byte}(\hat{b}_{c+k-1}, h_{c+k-1})$
            \State $\hat{b}_{c+k} \sim p(x)$
        \EndWhile
        \\
        \Return generated bytes $\tilde{b}_{1:c} + \hat{b}_{c+1:c+k}$, $M_\text{byte}$ last hidden state $h_{c+k}$, $M_\text{subword}$ hidden state $\hbar_{c'}$ as cache to restart from.
    \end{algorithmic}
    \caption{Speculative decoding iteration with subword drafter and byte-level verifier and corrector. (We use $\tilde{b}$ to indicate a drafted byte, while $\hat{b}$ denotes a corrected byte.)}
    \label{alg:spec}
\end{algorithm}



    \section{Synthetic noise settings}
\label{app:noise}

To confirm the robustness of \ssmname to input text corruptions, we employ the following synthetic noise settings adapted from \citet{xue2022byt5}: 
\begin{enumerate}[label=\arabic*)]
    \item \textit{Drop}: Bytes are dropped with a pre-set probability.
    \item \textit{Repetition}: Bytes are repeated one to three times (with equal likelihood).
    \item \textit{Antspeak}: Every character is capitalized and padded with spaces.
    \item \textit{Uppercase}: Characters are converted to uppercase with a pre-set probability.
    \item \textit{Random case}: Every character is set to a random case. 
\end{enumerate}

In addition to these five settings, we include the \textit{character swap} setting, where consecutive bytes are swapped with some probability.
    \section{PG19 generation samples}
\label{app:samples}

This section includes a few sample generations from the \mediumssm trained on the PG19 dataset. We use Nucleus sampling with $p = 0.98$ \citep{holtzman2020curious} and generate continuations for a total of $8,192$ bytes (including the given context prefix). Furthermore, we chose the same test set prefixes used in Appendix~$\text{F}$ of \citet{rae2020compressive}. We observe that the model is able to continue the dialogue in the style of the prefix and effectively recall the character names over hundreds of bytes.

\subsubsection*{Baby Mine by Margaret Mayo}

Context ($487$ bytes):
\sample{samples/prefixes/860.txt}

\mediumssm:
\sample{samples/mambabyte/860.txt}

\subsubsection*{The Diary of Samuel Pepys}

Context ($826$ bytes):
\sample{samples/prefixes/4128.txt}

\mediumssm:
\sample{samples/mambabyte/4128.txt}

\subsubsection*{The Patrol of the Sun Dance Trail by Ralph Connor}

Context ($1,059$ bytes):
\sample{samples/prefixes/3247.txt}

\mediumssm:
\sample{samples/mambabyte/3247.txt}

\end{appendices}

\end{document}